\documentclass[10pt]{article} 

\usepackage[accepted]{rlj} 
\usepackage{amssymb}    
\usepackage{mathtools}
\usepackage{mathrsfs}   
\usepackage{graphicx}  
\usepackage{subcaption}      
\usepackage[space]{grffile}  
\usepackage{url}             
\usepackage{lipsum}        

\title{Agent-centric learning: from external reward maximization to internal knowledge curation}
\setrunningtitle{Agent-centric empowerment}
\author{Hanqi Zhou\textsuperscript{1,2}, Fryderyk Mantiuk\textsuperscript{1}, David G. Nagy\textsuperscript{1,2}, Charley M. Wu\textsuperscript{1,2,3,4}}
\emails{hanqi.zhou@uni-tuebingen.de}

\affiliations{
$^{1}$\textbf{Human and Machine Cognition Lab, University of T{\"u}bingen, Germany}\\
$^{2}$\textbf{Department of Computational Neuroscience, Max Planck Institute for Biological Cybernetics, Germany}\\
$^{3}$\textbf{Centre for Cognitive Science, Institute of Psychology, Technical University of Darmstadt, Darmstadt, Germany}\\
$^{4}$\textbf{Hessian.AI, Darmstadt, Germany}
}


\begin{document}

\maketitle  

\begin{abstract}
    The pursuit of general intelligence has traditionally centered on external objectives: an agent's control over its \textit{environments} or mastery of specific \textit{tasks}. This external focus, however, can produce specialized agents that lack adaptability. 
    We propose \textit{representational empowerment}, a new perspective towards a truly agent-centric learning paradigm by moving the locus of control inward. This objective measures an agent's ability to controllably maintain and diversify its own knowledge structures. We posit that the capacity---to shape one's own understanding---is an element for achieving better ``preparedness'' distinct from direct environmental influence. 
    Focusing on internal representations as the main substrate for computing empowerment offers a new lens through which to design adaptable intelligent systems.
\end{abstract}

\section{The challenges of task- and environment-centric learning}
\label{sec:intro}
``Intelligence is what you use when you don’t know what to do.''
— Jean Piaget

Reinforcement Learning (RL) has made great progress in training agents to excel at narrow tasks by maximizing rewards \citep{sutton1998introduction}. Yet as termed by \citet{abel2024three}, its core dogmas---the reward hypothesis (all goals as reward maximization) and the environment spotlight (focus on modeling environments over agents)---reveal a tension. 
An agent optimized for a single task reward in a well-defined environment often struggles when new tasks not incentivized by its original training \citep{ringstrom2022reward, alet2020meta}. 
Given that agents, over a lifetime, will have to learn many aspects of the world, and since we cannot simulate all possible worlds for them to learn in, the current learning paradigm makes it hard to achieve broadly applicable intelligence---a high level of preparedness for unforeseen challenges.

Addressing the limitations suggests a new look beyond purely external task objectives and environmental designs. A promising direction involves a transition from an external task-centric and environment-centric viewpoint to a more internal agent-centric perspective \citep{singh2009rewards}. 
This agent-centric view prioritizes the development of internal representations that allow an agent to understand, adapt, and act effectively even when external objectives are novel or underspecified.  

The shift naturally raises a further question: by what \textit{principle} should these internal representations be managed to best prepare an agent for future challenges? 
The information-theoretic concept of \textit{empowerment} \citep{klyubin2005all, salge2014empowerment, lidayan2025intrinsically, mantiuk2025curiosity} offers a promising but underspecified framework. Empowerment quantifies an agent's potential to influence its future, often measured by the range and controllability of reachable states in the environment. This inherent link to having diverse options makes it a good candidate for formalizing ``preparedness''.
However, existing approaches to empowerment still focus on an agent's control over \textit{external environmental states}. Instead, a truly agent-centric view invites us to redirect this lens inward. If an agent's capacity is shaped by its internal knowledge, then its ability to control and diversify these internal representations could be a more direct route to robust adaptability.

Here, we offer a new perspective towards agent-centric learning. By applying empowerment to the agent's internal representations, instead of asking which state in the external environment one should reach, we ask \textbf{what kind of internal representational structures should an agent form and curate to maximize its ``preparedness'' for a diverse and unpredictable future?} 
This focus suggests that lasting adaptability may arise more profoundly from an agent's mastery over its own evolving understanding than from its immediate capacity to alter the external world. 

\subsection{Extrinsic reward maximization for task-centric learning} 
A key tenet of RL is that ``all of what we mean by goals and purposes can be well thought of as maximization of the expected value of the cumulative sum of a received scalar signal (reward)'' \citep{sutton1998introduction}. 
This statement, coupled with the view of intelligence as primarily goal achievement \citep{mccarthy1998spoken}, underpins the influential ``reward-is-enough'' hypothesis: that maximizing reward is a sufficient objective for developing general intelligence \citep{silver2021reward}. 

In this formulation, the reward signal is often a scalar feedback to guide the agent learning in an \textit{environment}, indicating the desirability of its actions (or states). 
Traditionally, the reward comes externally from the environment and is predefined by the designer, about a \textit{task} in their mind to be solved.
The agent's objective then becomes finding a solution (policy) that maximizes the reward.

However, this \textit{task-centric learning} (when rewards are tied to specific tasks) faces several problems. It is challenging to craft rewards that precisely capture intended goals without incentivizing undesirable ``reward hacking'' \citep[e.g., a cleaning robot hiding a mess instead of collecting it;][]{krakovna2020specification, skalse2022defining}. 
Even if the reward function is not correlated with any unintended objectives, it may lack the expressive power to represent all desirable orderings over policies or complex goal structures \citep{abel2021expressivity, bowling2023settling}. This can lead to ``steady-state type'' failures, where agents repeatedly attempt an impossible action (e.g., phasing through a wall) if that naively maximizes a flawed reward signal (e.g., given the goal of going to the next room).

Thus, while effective for achieving specific tasks in well-defined environments, this exclusive reliance on external reward maximization fails to guide the development of the \textit{generalizable internal representations} essential for long-term adaptability in open-ended problems \citep{hubinger2019risks}, such as autonomous robots in unpredictable settings or AI systems for creative discovery.

\subsection{Intrinsic motivation offers a band-aid fix towards environment-centric learning} 
The challenge of developing agents that can generalize beyond training tasks led to \textit{intrinsic motivation}---task-agnostic learning objectives that encourage exploration and skill acquisition. 
These internal rewards, generated by the agent itself, are typically based on principles like curiosity \citep[seeking novelty or surprise;][]{pathak2017curiosity, achiam2017surprise}, learning progress \citep[improving a model of the environment;][]{houthooft2016vime}, or competence \citep[achieving self-set goals;][]{colas2022autotelic}.

While intrinsic motivation has demonstrably improved learning about the current environment, it remains fundamentally \textit{environment-centric} in two key ways. 
First, its objectives are inherently tied to the external world. For example, novelty seeking encourages visiting all environmental states, and learning progress drives the formation of a more accurate transition model of the current environment \citep{modirshanechi2022taxonomy}. The focus, while broader than a single task, remains on ``what is out there to be known or done''. This can lead to overfitting to the specifics of the current environment. 
Second, intrinsic motivation functions by augmenting or replacing the external reward signal. The agent's learning still aims to maximize a (now potentially composite) scalar reward. While internal representations are learned and refined in this process, they are developed to the extent that they support this reward-seeking behavior, not necessarily because they possess inherent qualities, e.g., high compositionality, that would directly help future learning in different contexts. 

Thus, while intrinsic motivation pushes agents beyond myopic exploitation of external rewards and encourages more thorough engagement with their environment, it does not fully address the challenge of building internal cognitive structures designed for lasting adaptability \citep{abel2018state}. The resulting representations, though often richer, are still predominantly shaped by environmental regularities, which may not generalize to new settings.

\subsection{Empowerment as a general-purpose objective}
Given that existing reward design schemes (both extrinsic and intrinsic) remain tethered to predefined environments, how can an agent develop more general-purpose adaptability? 
One promising direction lies in identifying \textit{instrumentally convergent goals} \citep{bostrom2012superintelligent}---rewards that are beneficial across final goals in a wide range of environments and tasks \citep{omohundro2018basic}. 
Intuitively, for vastly different long-term objectives, from manufacturing paperclips to exploring galaxies, certain intermediate goals are consistently pursued not because they are inherently valuable, but because they are instrumental stepping stones. 
For example, any society is likely to develop highly efficient energy transportation, e.g., superconducting cables. This is not because conductivity is a terminal goal for society's values, but because energy is a powerful enabler for nearly any non-trivial goal. 

The concept of empowerment, first introduced by \citet{klyubin2005all}, offers an information-theoretic way to quantify such a general-purpose objective. Empowerment measures an agent's capacity to reliably bring about diverse futures regardless of the final goal. 
Formally, we can define the traditional, \textit{environment-centric empowerment} ($\operatorname{EnvEmp})$ as the degree of control an agent has over its external environment through the channel capacity (mutual information $I$) between its sequence of actions $a_{1:T}$ and the resulting environmental state $s'$ from its current state $s$:
\begin{equation}
\operatorname{EnvEmp}(s) = \max _{\pi\left(a_{1: T}\right)} I\left(s^{\prime} ; a_{1: T} \mid s\right)
\end{equation}
High empowerment signifies that the agent's actions can reliably lead to many distinct environmental states, a capacity that serves as a proxy for preparedness.
However, environmental empowerment, while a step towards general-purpose objectives, inherits some limitations previously discussed. 
By focusing on control over externally defined environmental states $s$ and $s'$, it remains susceptible to the ``frame-of-reference'' problem \citep{clancey2014frame}. If the definition of these states is imposed externally or is not learnable by the agent, then maximizing environmental empowerment might still lead to specialization to the idiosyncrasies of that particular representation \citep{mantiuk2025curiosity}.

This brings us to the central thesis of this paper: 
General intelligence and adaptability may require shifting the locus of empowerment from the external environment to the agent's own \textit{internal representations}. The critical capacity might not just be to control the world, but to control, adapt, and expand the very way the agent models and understands the world.

\section{Agent-centric representational empowerment}
\subsection{From environmental to representational states}
Environmental empowerment has been shown to explain human behavior that we value the potential of diverse options \citep{brandle2023empowerment, du2023can}, although it is subject to the environmental dynamics \citep{mantiuk2025curiosity}.
Here, we define \textbf{agent-centric empowerment} on agent's internal representational structures that maximizes its ``preparedness'', which in this sense, refers to its capacity to generate or reconfigure its knowledge to effectively address unforeseen tasks.

Consider a Minecraft world \citep{hafner2021benchmarking}, where an agent driven by \textit{environmental empowerment} aims to maximize its control over immediate surroundings. This may translate to building a wooden fortress optimized for the local terrain in a forest.
However, this specialization becomes suboptimal if the agent moves to a barren desert.
In contrast, an agent driven by \textit{representational empowerment} focuses on the knowledge of building techniques and material properties, instead of investing all resources into one perfect base. When the challenge changes to woodland, skyblock or trial chambers, it can craft new shelters or tools from available items---without needing prior exposure.

Thus, rather than entirely re-defining empowerment, we re-frame the pertinent states involved. Our formulation follows the spirit of the ``bitter lesson'' \citep{sutton2019bitter}: instead of relying on fixed human-specified knowledge, we propose that agents explicitly learn to build their own knowledge by making the dynamics of representation learning the primary target of optimization.

The Minecraft analogy can generalize to abstract knowledge and naturally incorporates the notion of resource constraints \citep{lieder2020resource}. 
An agent operates under both memory costs, associated with storing its knowledge library $Z_k$, and computational costs. Representational empowerment helps mediate the trade-off between them. Computational costs can be understood as twofold: the ``offline'' effort required for the curator to build and maintain an empowered library, and the ``online'' effort for the executor to adapt that library to solve a new task. By investing more upfront in offline computation to build a library (at a certain memory cost), the agent can amortize future learning, reducing the online computation needed to solve subsequent problems.


\subsection{Potential of diverse and controllable representations internally}
Let us assume the agent learns over a sequence of tasks. Each task $\tau_k \in \{\tau_1, \tau_2, \ldots\}$ is a Markov Decision Process (MDP) $(\mathcal{S}_k, \mathcal{A}_k, \mathcal{P}_k, R_k)$, with environmental states $\mathcal{S}_k$ (e.g., the observed blocks) and actions $\mathcal{A}_k$ (e.g., chopping a tree). 
The agent maintains an internal library of representations, $Z_k$ (e.g., design of diamond axe), accumulated from prior tasks $\tau_1, \ldots, \tau_k$.  
Upon engaging with task $\tau_{k+1}$, it may acquire a new piece of knowledge $\hat{Z}_{k+1}$ (e.g., construction of bamboo planks). 

Here we use meta reinforcement learning \citep{botvinick2019reinforcement} to explain the two components in learning:
1) A meta-level \textit{curator} responsible for evolving the internal representational library $Z_k$, learned from past tasks $\tau_{1:k}$ to maximize its representational empowerment.
2) A task-level \textit{executor} that uses the curated representation $Z_k$ to find solutions for the following specific task $\tau_{k+1}$.

\subsubsection{Curator: representational empowerment maximization}
We frame the curator's decision-making at the meta level. At each step $k$, the curator observes a state $s_k^c = (Z_{k-1}, \hat{Z}_k)$---its current library and new knowledge---and selects an integration action $a_k^c \in \mathcal{A}^c$. 
These actions, e.g., selecting, composing, or pruning, produce the next library, $Z_k = a_k^c(Z_{k-1}, \hat{Z}_k)$. The curator's goal is to maximize an intrinsic reward $r_k^c(Z_k) = \operatorname{RepEmp}(Z_k)$, the \textit{representational empowerment} of the resulting library $Z_k$ (defined below in Eq.~\ref{eq:rep_empowerment}).

This reward $\operatorname{RepEmp}(Z_k)$ is calculated via multiple simulated roll-outs. The agent imagines applying a sequence of \textbf{modification operations}, $\omega_k^{1:T} = \{\omega^1_k, \ldots, \omega^T_k\}$ drawn from a set of available operations $\Omega$, to $Z_k$. Using $\omega_k^{1:T}=\boldsymbol\omega_k$ for simplicity, this yields a modified library $Z'_k \sim p(Z'_k \mid Z_k, \omega_k)$. Then $\operatorname{RepEmp}(Z_k)$ is the channel capacity between these imagined operations and their outcomes, quantifying control over the agent's own representational state:
\begin{equation}
    \operatorname{RepEmp}(Z_k) = \max_{\boldsymbol\omega_k \in \Omega^T} I(Z'_k ; \boldsymbol\omega_k \mid Z_k)
    \label{eq:rep_empowerment}
\end{equation}
Here, $\boldsymbol\omega_k$ is sampled from the $\Omega^T$, $T$-fold Cartesian product of $\Omega$, and denotes an operation sequence over $T$ time steps. A high empowerment value, resulting as the reward for action $a_k^c$, means the library $Z_k$ is both diverse and controllable. The horizon $T$ also reflects a computational budget for how much $Z_k$ can be internally modified by the curator, or by the executor to adapt it later.

We aim to distinguish the high-level actions ($a^c$), on how to update the library globally, from primitive modification operations ($\omega$). Operations $\omega$ are fine-grained transformations to update pieces of knowledge representation, e.g., the continuous interpolation of high-dimensional features, or symbolic rules (e.g., mutation) for discrete modules, from $Z_{k-1}$ and $\hat{Z}_k$.

Decomposing $\operatorname{RepEmp}(Z_k)$, we can see better that it encourages libraries $Z_k$ that have the potential for diversity (i.e., can be modified into many distinct forms $Z'_k$) and controllability (i.e., the transformation cannot be arbitrary):
\begin{equation}
    I(Z'_k ; \omega_k \mid Z_k) = H(Z'_k \mid Z_k) - H(Z'_k \mid Z_k, \omega_k)
    \label{eq:mi_decomposition}
\end{equation}
The first term, $H(Z'_k \mid Z_k)$, quantifies the \textbf{diversity} of representation $Z'_k$ reachable from $Z_k$. Higher diversity suggests $Z_k$ can be transformed into a wide range of different representations, potentially useful for an as-yet-unknown task $\tau_{n+1}$.
The second term, $H(Z'_k \mid Z_k, \omega_k)$, quantifies the average \textbf{uncertainty} of the outcome $Z'_k$ given a sequence of operations $\omega_k$. A lower value implies that the operations have more predictable effects, enabling precise control. This term favors representations that are not only broadly transformable but also have controllable evolutions.

\subsubsection{Executor: task-specific adaptation} 
Once the curator establishes an empowered representation $Z_k$, the executor uses it for the next task $\tau_{k+1}$ in two potentially intertwined phases:

\textbf{Representation tuning}: the executor may first apply a bounded sequence of operations $\omega_k^{1:T'}$ to mold the generic library into a task‑tailored variant $Z_k^{\star}$. Because $Z_k$ was optimized for high adaptability, a short horizon $T'$ may suffice to reach a configuration beneficial for the new task.

\textbf{Task completion}: Using $Z_k^{\star}$ (or $Z_k$ directly), the executor interacts with the environment to maximize extrinsic reward $R_{k+1}$. 
The executor can also interleave further operations (fine-tuning $Z_k^{\star}$) with policy updates. For example, if a particular skill from $Z_k^{\star}$ is almost effective but needs slight adjustments, the executor can adjust it. This creates a \emph{use-improve} cycle: observed task performance provides feedback on which representational refinements are most beneficial, turning the curator's long-term investment into improved sample efficiency and asymptotic performance.

\section{Example: empowerment through curating the program library}
The representations $Z$ are preferably symbolic (e.g., programs, objects) to support interpretable representational operations, such as abstraction, composition \citep{rule2024symbolic, ellis2021dreamcoder, zhou2024harmonizing}. 
We provide an example of how to instantiate representational empowerment through symbolic programs (Sec.~\ref{sec:melody-program}), which offer several advantages as a representational format \citep[e.g., compositionality and generalization;][]{lake2015human, rule2020child}. 

Formally, we can define a space of program representations $\mathcal{L}$ where each program $z \in \mathcal{L}$ could be a causal model \citep{icard2017programs}, policies \citep{correa2025exploring}, values, or goals \citep{davidson2025goals} forming the cognitive structure.
For generality, each program contains a function term and parameters, e.g., \texttt{play(instrument)} has the function \texttt{play} and can have parameter \texttt{violin} which is already highly abstract with typed parameters (type \texttt{instrument}).  
Drawing inspiration from evolutionary algorithms and genetic programming \citep{forsyth1981beagle}, the representational operations $\Omega$ can include: \textbf{selection} for saving effective programs, function-level \textbf{crossover} for combining fragments from multiple programs to create new ones \citep{o2009fragment}, \textbf{abstraction} for creating higher-level programs \citep{bowers2023top}, and parameter-level \textbf{mutation} for modifying parameters of existing programs \citep{franken2022algorithms}. 

\subsection{Learning melodic programs} 
\label{sec:melody-program}
Consider an agent facing a sequence of tasks $\{\tau_1, \tau_2, \ldots\}$. Each task $\tau_k$ requires memorizing and playing a specific target melody, $M_k^{\text{target}}$. For a task $\tau_k$, the executor uses the current library $Z_{k-1}$ to match the melody $M_k^{\text{target}}$. Its actions can be primitive (e.g., \texttt{add\_note(C4)}) or executing a program (e.g., \texttt{repeat(C4, 2)}). The executor receives a reward, $R_k$, based on the similarity between its generated melody and the target. An empowered library $Z_{k-1}$ would enable generation more efficiently than \texttt{add\_note} verbatim. 
When finishing $\tau_k$, the curator might get a new melodic fragment, $\hat{Z}_k$. This new program is a candidate for the library. The curator then selects an integration action $a_k^c$ to produce the new library $Z_k = a_k^c(Z_{k-1}, \hat{Z}_k)$ guided by maximizing $\operatorname{RepEmp}(Z_k)$.

\textbf{Programs evolution. }
After exposure to some tasks (e.g., playing simple folk melodies), the library $Z_{k-1}$ contains:
\texttt{up/down(n, steps)}(increases/decreases the pitch), 
\texttt{repeat(pattern, times)}.
In a new task $\tau_k$ (e.g., playing chordal harmony and arpeggios), the agent learns two new programs $\hat{Z}_k$, \texttt{arpeggio(root, chord, direction)} and \texttt{sequence(note, pattern)}.
\texttt{arpeggio} generates an arpeggio starting from \texttt{root}, using notes from the \texttt{chord} type (e.g., major, minor).
\texttt{sequence} generates notes starting from \texttt{note}, following the \texttt{pattern}. 

The agent can apply modification operations on these programs ($Z_{k-1}$ and $\hat{Z}_k$). 
For example, the agent recognizes that \texttt{up} and \texttt{down} are special cases of \texttt{move(direction, n, steps)} with direction so it \textbf{abstracts} over them. 
The agent can combine, applying \textbf{crossover} over \texttt{arpeggio} and \texttt{repeat} to create $\texttt{repeated\_arpeggio(root, chord, direction, times)}$.

\subsection{Curating a library with regularized diversity}
The agent curates its library $Z_k$ based on representational empowerment (Eq.~\ref{eq:rep_empowerment}). Rather than maximizing raw diversity ($H(Z'_k \mid Z_k)$ in Eq.~\ref{eq:mi_decomposition}), the process balances two key principles: task relevance, by integrating useful new programs ($\hat{Z}_k$), and controllable adaptability, by penalizing transformational uncertainty ($H(Z'_k \mid Z_k, \omega_k)$).

\textbf{Task relevance as a filter. }
A newly synthesized program $\hat{z} \in \hat{Z}_k$ (e.g., \texttt{arpeggio} derived from task $\tau_5$) is considered for long-term integration into $Z_{k-1}$, because it has proved usefulness within the task $\tau_k$. This task performance acts as an initial, pragmatic filter. 

\textbf{Avoiding a single nearly universal program. }
This term $H(Z'_k \mid Z_k, \omega_k)$ in Eq.~\ref{eq:mi_decomposition}, which is subtracted, quantifies the average \textit{ambiguity or lack of precision}.
Consider a representation $Z_k$ that is overly flexible, if a program like \texttt{generate\_any\_melody(latent)} exists. It could be a very large, unconstrained, but well-trained neural decoder mapping from a latent space to represent melodies. 
Here, typical operations (e.g., \texttt{mutation} for small parameter perturbations) result in highly wild and unpredictable changes to the melody produced. While it might theoretically be able to produce any melody, $p(Z'_k \mid Z_k, \omega_k)$ is diffuse and high, and the problem for the agent becomes finding a way to choose from the latent space, in order to reliably arrive at the desired melody.
Such a representation can prevent the agent from effectively ``sculpting'' useful representations. 

Potentially, after the agent evaluates which representations provide the most empowerment, it decides to keep \texttt{move}, \texttt{arpeggio}, but discards the now-redundant \texttt{up} and \texttt{down}.

\subsection{Comparing program libraries}
We further illustrate how empowerment guides this by comparing potential libraries. For simplicity, we assume two available representational operations $\Omega$: \textbf{crossover} and \textbf{mutation}. 

\textbf{Diversity preference. }
Assume two melodic library candidates: $Z_A = \{\texttt{up}, \texttt{down}\}$ and $Z_B = \{\texttt{move}, \texttt{repeat}\}$ (programs explained in Sec.~\ref{sec:melody-program}). Each program can be applied with \textbf{crossover} into $M=3$ distinct variants by converting its style, rhythm, or articulation (e.g., \texttt{up\_staccato}, \texttt{move\_smooth}, \texttt{repeat\_accelerando}).

Modifications for the 2 programs in library $Z_A$ can lead to $M^2 = 9$ distinct libraries $Z'_A$ (e.g., $\{\texttt{up\_staccato}, \texttt{down}, \texttt{repeat}\}$).
Though they are syntactically different, the functional diversity might be less. Crossovers of \texttt{up} and \texttt{down} overlap because of the music octave. 
We estimate that the $9$ variants from \texttt{up} and \texttt{down} together yield approximately $M+\delta \approx 6$ effectively distinct libraries, $N_{\text{eff}}(Z_A) \approx 6$.
So, $\operatorname{RepEmp}(Z_A) = H(Z'_A \mid Z_A) \approx \log_2(6) \approx 2.59$ bits.

The library $Z_B$ also has $9$ syntactically distinct libraries $Z'_B$.
Here \texttt{move} provides richer functions because \textbf{mutation} can create higher-order structures considering parameter \texttt{direction}, e.g., \texttt{move\_staccato\_rhythmic(rhythm\_X, steps)} (applies a rhythmic pattern to the movement while having staccato). 
These $3$ variants of \texttt{move} could provide at least $2\times M=6$ distinct functional capabilities. 
Thus, $N_{\text{eff}}(Z_B)$ is estimated as $6 \times 3 = 18$.
So, $\operatorname{RepEmp}(Z_B) = H(Z'_B \mid Z_B) \approx \log_2(18) \approx 4.17$ bits.

\textbf{Controllability preference. }
Assume a new library candidate $Z_C$ = \{\texttt{neural\_gen(latent)}, \texttt{repeat}\}. Here, \texttt{neural\_gen} is a powerful neural melody generator. 
For \texttt{neural\_gen}, suppose there are $20$ mutations (the size of the latent space for its parameter). Among them, outcomes of $5$ latents are predictable ($\omega^{\texttt{pred}}$) and unique. 
outcomes of $15$ latents are unstable ($\omega^{\text{unstable}}$), having equal chance of producing `style alpha` and `style beta`. 
The diversity of $N_{\text{eff}}(Z_B) \approx 7 \times 3 = 21$, and $H(Z'_C \mid Z_C) \approx \log_2(21) \approx 4.39$ bits which reflects high potential. 
There is a punishment for the uncertainty that the policy considers $20\times 3=60$ combinations, thus $H(Z'_C \mid Z_C, \omega_C) = \frac{(5 \times 3 \times  0) + (15 \times 3 \times \log2 \text{ bit})}{60} = \frac{15}{23} \approx 0.75$ bits.
So, $\operatorname{RepEmp}(Z_C) \approx 4.39 - 0.75 = 3.64$ bits.

\textbf{Decision and interpretation. }
The library $Z_B$ with the more abstract program (\texttt{move}) is more empowered, $\operatorname{RepEmp}(Z_B)$ > $\operatorname{RepEmp}(Z_A)$ ($4.17 > 2.59$), because its components can be transformed into a more functionally diverse set. 
The diversity of $Z_C$($4.39$) is actually higher than $Z_B$ ($4.17$) because of a powerful general program \texttt{neural\_gen}. However, when considering the penalization of uncontrollable mutations, i.e., those programs which require another search over parameter space, $Z_B$ is preferred.

\section{Discussions}
We propose an agent-centric learning paradigm, based on \textit{representational empowerment}, in which an agent maximizes its capacity to controllably diversify its internal knowledge, rather than the external world. This framework offers a new direction for building adaptable agents while also extending key ideas from AI, cognitive science, and evolutionary theory.

\textbf{Knowledge cultivation in AI. }
From a continual learning perspective, an agent's internal library $Z_k$ is its evolving knowledge. 
Unlike standard Bayesian updating (e.g., Bayes-Adaptive MDPs maintaining beliefs over MDPs) that aims to assimilate \textit{all} new evidence $\hat{Z}_k$, representational empowerment guides a \textit{selective} curation of $Z_k$ rather than updating beliefs based on a fixed parameterization of past experiences \citep{bowling2025rethinking}. The goal of curating a library for fast adaptation is also conceptually related to meta-learning, e.g., Model-Agnostic Meta-Learning (MAML) \citep{finn2017model}. MAML learns a parameter initialization that can be quickly fine-tuned to new tasks via gradient descent. However, MAML's meta-objective is only tied to performance on a distribution of tasks. Representational empowerment, in addition, uses an intrinsic objective---the empowerment of the library itself---to foster task- and environment-agnostic adaptability.

That said, operationalizing $\operatorname{RepEmp}(Z_k)$ (Eq.~\ref{eq:rep_empowerment}) presents both key challenges but also promising future directions. 
First, the framework's effectiveness is sensitive to the (pseudo-)metric or kernel used to measure distance in the representation space, which is important for computing entropy terms $H(Z'_k \mid Z_k)$ and $H(Z'_k \mid Z_k, \omega_k)$. A purely syntactic metric might be brittle. 
A more robust approach could define a \textit{functional} metric, where the ``distance'' between two representations is measured by the behavioral difference they produce when executed. This metric could even be learned, for instance through contrastive methods, to capture a task-relevant notion of similarity. 

Second, representational empowerment is sensitive to the predefined set of modification operations, $\Omega$. One direction is to treat the set of operations not as fixed, but as a dynamic, learnable component. 
An agent could start with a set of primitive, cognitively-inspired operations (e.g., copy, compose, abstract) and learn to construct more powerful macro-operations over time. This turns $\Omega$ into a second-order library that co-evolves with the primary knowledge library $Z_k$, creating a virtuous cycle of cognitive growth and aligning with the idea of a shared ``cognitive toolkit'' discussed below.

\textbf{Information processing in cognitive science. }
Aligning with the perspective of resource rationality in cognitive science \citep{lieder2020resource}, representational empowerment exchanges computational costs with memory costs brought by the library $Z_k$, which amortizes the learning effort. 
This extends with classical information-theoretic objectives that can be agnostic to the cost of deriving or searching for representations. For instance, frameworks like the Information Bottleneck \citep{tishby2000information}, which compresses an input $X$ to preserve information about a \textit{specific target} $Y$, or Predictive Information \citep{bialek2001predictability}, which captures past-future regularities, typically compress representations for a predefined data stream for ``memory'' regardless of any ``computation''. 
They are fundamentally about the efficient processing of sensory information. Here, we argue that an agent's long-term cognitive effectiveness may be better understood not just by how efficiently it processes information about its environment, but by the adaptive potential it cultivates within its own representational system.

\textbf{Socio-cultural dynamics. }
In a multi-agent context, the set of representational operations $\Omega$ can become a shared ``cognitive toolkit''. 
Though agents may share the same physical world, their different goals, values, and learned world models mean that the ``effective task'' each agent faces and the resulting representations are often unique \citep{molinaro2023goal, witt2024flexible}. 
Two agents in the same physical space can have divergent policies and interpretations of environmental affordances. 
Consequently, directly transferring policies or internal representations from one agent to another is often difficult, while the ability to exchange or co-develop these \textit{operations} is crucial for collective intelligence, just as human progress leverages shared conceptual tools \citep[e.g., language, mathematics;][]{wu2024group}. For example, effective education often focuses on problem-solving methodologies (e.g., mathematical proof techniques like induction) rather than just rote memorization of solutions to specific problems. 
Future work could explore how a population can collectively discover and disseminate operations that enhance their collective representational empowerment, forming a ``cultural ratchet'' for cognitive tools \citep{tennie2009ratcheting}.

\appendix

\subsubsection*{Acknowledgments}
\label{sec:ack}
We thank Alison Gopnik, Eunice Yiu and Fei Dai for helpful discussions.
The authors thank the International Max Planck Research School for Intelligent Systems (IMPRS-IS) for supporting HZ. 
HZ, DGN, \& CMW are supported by the German Federal Ministry of Education and Research (BMBF): Tübingen AI Center, FKZ: 01IS18039A, funded by the Deutsche Forschungsgemeinschaft (DFG, German Research Foundation) under Germany’s Excellence Strategy–EXC2064/1–390727645, and funded by the DFG under Germany's Excellence Strategy – EXC 2117 – 422037984.

\bibliography{main}
\bibliographystyle{rlj}
\end{document}